\documentclass[titlepage,12pt,letterpaper,oneside]{article}
\usepackage{epsf}
\usepackage{psfig}
\usepackage{times}
\usepackage{graphicx}
\psfigurepath{./pic_new/}

\newcommand{\R}{\mbox{I} \! \mbox{R} }

\newcommand{\algodiff}{0.5cm}
\newcommand{\algobreite}{15.5cm}

\renewcommand{\vec}[1]{{\bf #1}}
 
\title{\bf 
  A Computational Framework for \\
  Nonlinear Dimensionality \\
  Reduction of Large Data Sets: \\
  The Exploratory Inspection Machine (XIM)
  \\
  \vspace{1.0cm}
  \large Technical Report \\
  \vspace{0.5cm}
  \large Version 1.0 \\
  \vspace{0.5cm}
} 
  
\author{\large Axel Wism{\"u}ller, M.D., Ph.D. \\
\\
  \vspace{-0.1cm} 
  \normalsize Director, Computational Radiology Laboratory \\
  \vspace{-0.1cm} 
  \normalsize Associate Professor of Radiology \\
  \vspace{-0.1cm}
  \normalsize Associate Professor of Biomedical Engineering\\
  \vspace{-0.1cm}  
  \normalsize Dept. of Radiology and Dept. of Biomedical Engineering \\
  \vspace{-0.1cm}  
  \normalsize University of Rochester, New York \\
  \vspace{-0.1cm}
  \normalsize 601 Elmwood Avenue \\
  \vspace{-0.1cm}
  \normalsize Rochester, NY 14642-8648, U.S.A. \\
  \vspace{-0.1cm}
  \normalsize Phone: (585) 613-2399 \\
  \vspace{-0.1cm}
  \normalsize axel{\_}wismueller@urmc.rochester.edu
 \\
  \normalsize e-mail: axel@wismueller.de
}
 
\begin{document}
 
\maketitle

\setlength{\baselineskip}{1.7em}
 
\thispagestyle{empty}
\section*{Abstract}
\noindent In this paper, we present a novel computational framework for nonlinear dimensionality reduction which is specifically suited to process large data sets: the Exploratory Inspection Machine (XIM). XIM introduces a conceptual crosslink between hitherto separate domains of machine learning, namely topographic vector quantization and divergence-based neigbor embedding approaches. There are three ways to conceptualize XIM, namely (i) as the inversion of the Exploratory Observation Machine (XOM) and its variants, such as Neighbor Embedding XOM (NE-XOM), (ii) as a powerful optimization scheme for divergence-based neighbor embedding cost functions inspired by Stochastic Neighbor Embedding (SNE) and its variants, such as t-distributed SNE (t-SNE), and (iii) as an extension of topographic vector quantization methods, such as the Self-Organizing Map (SOM). By preserving both global and local data structure, XIM combines the virtues of classical and advanced recent embedding methods. It permits direct visualization of large data collections without the need for prior data reduction. Finally, XIM can contribute to many application domains of data analysis and visualization important throughout the sciences and engineering, such as 
pattern matching, constrained incremental learning, data clustering, and the analysis of non-metric dissimilarity data.
 
\newpage
\setcounter{page}{1}

\section{Motivation}

In this paper, we present a novel computational framework for nonlinear dimensionality reduction which is specifically suited to process large data sets: the Exploratory Inspection Machine (XIM). The central idea of XIM is to invert our recently proposed variant of the Exploratory Observation Machine (XOM)~\cite{Axel2006_Diss}, \cite{Axel_wsom_2009_embedding}, \cite{Axel2009_esann_embedding}, namely the Neighbor Embedding XOM (NE-XOM) algorithm as proposed in~\cite{Bunte2011}
by systematically exchanging the roles of ordering and exploration spaces in topographic vector quantization mappings. In contrast to NE-XOM, this is accomplished by calculating the derivatives of divergence-based topographic vector quantization cost functions {\em with respect to distance measures in the high-dimensional data space} rather than derivatives {\em with respect to distance measures in the the low-dimensional embedding space.} 

As an important consequence, we can thus introduce successful approaches to alleviate the so-called 'crowding phenomenon' as described by~\cite{van_der_Maaten2008} into classical topographic vector quantization schemes, in which, in contrast to XOM and its variants, data items are associated with the {\em exploration} space and not with the {\em ordering} space of topographic mappings. The resulting learning rules for heavy-tailed neighborhood functions in the low-dimensional embedding space are significantly different from corresponding update rules for NE-XOM and its variants for other divergences.

XIM extends classical topographic vector quantization methods in that its cost function does not only optimize `trustworthiness' in the sense of~\cite{Venna2007}, i.e. small distances in the embedding space correspond to small distances in the data space: By introducing repulsive forces into the learning rules, they optimize `continuity' as well, such that small distances in the data space are represented by small distances in the embedding space. This is exactly opposite to NE-XOM, in which divergence-based cost functions introduce repulsive forces that help to optimize `trustworthiness', whereas the original XOM cost function predominantly optimizes `continuity'.

In the light of these considerations, there are three ways to conceptualize XIM as a general computational framework for nonlinear dimensionality reduction, namely (i) as the systematic inversion of the Exploratory Observation Machine (XOM)~\cite{Axel2006_Diss}, \cite{Axel2009_esann_embedding} and its variants, such as Neighbor Embedding XOM (NE-XOM)~\cite{Bunte2011}, (ii) as a powerful optimization scheme for divergence-based dimensionality reduction cost functions inspired by Stochastic Neighbor Embedding (SNE)~\cite{Hinton2003} and its variants, such as t-distributed SNE (t-SNE)~\cite{van_der_Maaten2008}, and (iii) as an extension of topographic vector quantization methods, such as the Self-Organizing Map (SOM)~\cite{Kohonen2001}, where XIM introduces the option to explicitly optimize `continuity'. Thus, XIM introduces novel conceptual crosslinks between hitherto separate domains of machine learning. 

In the remainder of this paper, we first motivate XIM in the context of topographic vector quantization. We then derive its learning rule from a cost function which is based on the generalized Kullback-Leibler divergence between neighborhood functions in the data and embedding spaces. We further introduce the t-distributed and Cauchy-Lorentz distributed XIM (t-XIM and c-XIM) variants which help to address the crowding phenomenon in nonlinear dimensionality reduction. After specifying technical details of how to calculate the XIM learning rule by computing the derivative of its cost function, we discuss algorithmic properties and variants of XIM, including batch computation and XIM for non-metric data. After presenting experimental results, we finally extend XIM beyond the Kullback-Leibler divergence by deriving XIM learning rules for various other divergence measures known from the mathematical literature.

\section{Topographic Mapping}

We first motivate the Exploratory Inspection Machine (XIM) in the context of topographic vector quantization techniques. In their simplest form, these algorithms may be seen as methods that map a finite number $N$ of data points $\vec{x}_i \in \R^D, i \in \{1,\ldots,N\}$ in an exploration space $E$ to low-dimensional image points $\vec{y}_i \in \R^d$ in an ordering space $O$. The assignment is $\vec{x}_i \mapsto \vec{y}_i$ and typically $d \ll D$, e.g.~$d=2,3$ for visualization purposes. The mapping $\vec{x}_i \mapsto \vec{y}_i$ is not calculated directly, rather the data points $\vec{x}_i$ are represented by so-called `prototypes' $\vec{w}_j \in E \subset \R^D, j \in \{1,\ldots,M\}$ that are mapped to target points $\vec{r}_j \in O \subset \R^d$, so-called `nodes'. These nodes are priorly chosen as a structure hypothesis in the observation space $O$. Reasonable choices are the location of nodes on a regular lattice structure, e.g.~a  rectangular or hexagonal grid, but no specific limitations apply, i.e. arbitrary geometrical and even random arrangements may be considered.

Like in topographic vector quantizers, the goal of XIM is to find prototype positions $\vec{w}_j$ in such a way that the tuples $\left( \vec{w}_j, \vec{r}_j \right) \in E \times O$ represent pairs of reference points for defining a mapping $\Psi$ which allows the user to find a low-dimensional target point $\vec{y}_i$ for each data point $\vec{x}_i$. Such an explicit mapping based on computed pairs of reference points can be accomplished by various additional interpolation or approximation procedures, as shown later. 

Each prototype $\vec{w}_j$ defines a `receptive field' by a decomposition of $E$ according to some optimality rule for mapping points to prototypes, e.g.~by a criterion of minimal distance to a specific prototype. Let the spaces $E$ and $O$ be equipped with distance measures $d_E(\vec{x}, \vec{w}_j)$ and $d_O(\vec{r}_k, \vec{r}_l)$. 
A simple way to define a best-match node $\vec{r}^*$ for a given data vector $\vec{x}$ is to compute
\begin{equation}
	\vec{r}^*(\vec{x}) = \vec{r}_{\phi(\vec{x})} = \Psi(\vec{w}_{\phi(\vec{x})}),
\end{equation}
where $\phi(\vec{x})$ is determined according to the minimal distance criterion
\begin{equation}
	d_E(\vec{x}, \vec{w}_{\phi(\vec{x})}) = \min_j \left( d_E(\vec{x}, \vec{w}_j) \right).
\label{eq_minimal_distance}
\end{equation}

In a simple iterative approach, learning of prototypes $\vec{w}_j$ is accomplished by the online adaptation rule
\begin{equation}
 \vec{w}_j \leftarrow \vec{w}_j - \epsilon \, h_\sigma(d_O(\vec{r}^*,\vec{r}_j))\frac{\partial d_E(\vec{x}, \vec{w}_j)}{\partial \vec{w}_j}
\label{eq_xim_update}
\end{equation}
where $\epsilon > 0$ denotes a learning rate, $d_O$ and $d_E$ distances in the ordering and exploration spaces, respectively, e.g.~squared Euclidean, and $h_\sigma(t)$ a neighborhood cooperativity, which is typically chosen as a monotonically decreasing function, e.g. a Gaussian
\begin{equation}
	h_\sigma(t)=  \exp{\left(-\frac{t}{2\sigma^2}\right)}, \; \sigma > 0.
\end{equation}

However, it should be noted that the distance measures $d_E(\vec{x}, \vec{w}_j)$ and $d_O(\vec{r}_k, \vec{r}_l)$ are not confined to squared Euclidean: arbitrary, even non-metric distance measures may be applied. Likewise, the neighborhood cooperativity $h$ is not restricted to Gaussians: other choices, such as heavy-tailed distributions may provide advantages, as discussed below. 
   
\section{The Exploratory Inspection Machine (XIM)}

It can be shown that for a continuous distribution of inputs, the learning rule (\ref{eq_xim_update}) cannot be expressed as the derivative of a cost function~\cite{Erwin92b}. According to~\cite{Heskes99} one can circumvent this problem by replacing the minimal distance criterion (\ref{eq_minimal_distance}) by adopting a modified best-match node definition
\begin{equation}
	\vec{r}^*(\vec{x}) = \vec{r}_{\phi(\vec{x})} = \Psi(\vec{w}_{\phi(\vec{x})}),
  \label{eq_bestmatch}
\end{equation}

\noindent where $\phi(\vec{x})$ is determined according to
\begin{equation}
	\sum_j h_\sigma(d_O(\vec{r}_{\phi(\vec{x})},\vec{r}_j)) \, d_E(\vec{x}, \vec{w}_j) 
	= \min_k \left( \sum_j h_\sigma(d_O(\vec{r}_k,\vec{r}_j)) \, d_E(\vec{x}, \vec{w}_j) \right)
\label{eq_bestmatch_definition}
\end{equation}

\noindent This leads to the cost function

\begin{equation}
	E^{\prime} \propto \int \sum_i \delta_{\vec{r}^*(\vec{x}), \vec{r}_i} \sum_{j=1}^M h_\sigma(d_O(\vec{r}^*(\vec{x}),\vec{r}_j)) \, d_E(\vec{x}, \vec{w}_j) p(\vec{x}) d\vec{x},
\label{eq_xim_cost}
\end{equation}

\noindent where $\delta$ denotes the Kronecker delta. The derivative of (\ref{eq_xim_cost}) with respect to $\vec{w}_j$ using the best-match definition (\ref{eq_bestmatch_definition}) yields the learning rule (\ref{eq_xim_update}).

Given these settings, we propose to augment the cost function (\ref{eq_xim_cost}) by an additional term in order to combine fast sequential online learning known from topographic mapping, such as in learing rule (\ref{eq_xim_update}), and principled direct divergence optimization approaches, such as in SNE~\cite{Hinton2003}.
%
%
To this end, by means of cost function (\ref{eq_xim_cost}) we can define new learning rules based on the generalized Kullback-Leibler divergence for not normalized positive measures $p$ and $q$ with $0 \leq p,q \leq 1$
\begin{equation}
	E_{\mbox{\scriptsize GKL}}(p||q) = \int{[p(\vec{x}) \log \frac{p(\vec{x})}{q(\vec{x}})]} d\vec{x} - \int{[p(\vec{x}) - q(\vec{x})]} d\vec{x}
\end{equation}

We use the cooperativity functions $h_\sigma(d_O(\vec{r}_i, \vec{r}_j))$ and $g_\gamma(d_E(\vec{x}, \vec{w}_j))$ as positive measures, in the following abbreviated by $h_\sigma^{ij}$ and $g_\gamma^{j}$.

Inspired by t-SNE~\cite{van_der_Maaten2008}, neighborhood functions $h_\sigma^{ij}$ of the low-dimensional embedding space $O$ can be chosen as a heavy-tailed distribution, e.g. a Student $t$ or Cauchy-Lorentzian, in order to alleviate the so-called `crowding problem' which usually occurs in high dimensions related to the curse of dimensionality. It is important to notice that in XIM the neighborhood function of the embedding space refers to the {\em ordering space} $O$ of the topographic mapping. This marks a fundamental difference when compared to XOM~\cite{Axel2009_spie_embedding} and its variant NE-XOM~\cite{Bunte2011}, in which the neighborhood function of the embedding space refers to the {\em exploration space} $E$ of the topographic mapping.

Based on these settings, we obtain a novel cost function
\begin{equation}
	E_{\mbox{\scriptsize XIM}} \propto \int \sum_i \delta_{\vec{r}^*_{\mbox{\tiny GKL}}(\vec{x}), \vec{r}_i} \sum_j \left[ h_\sigma^{ij} \ln \left( \frac{h_\sigma^{ij}}{g_\gamma^{j}} \right) - h_\sigma^{ij} + g_\gamma^{j} \right] p(\vec{x}) d\vec{x}
\label{eq_cost_nexim}	
\end{equation}

\noindent where the best-match node $\vec{r}^*(\vec{x})$ for data point $\vec{x}$ is defined such that 
\begin{equation}
  \sum_j \left[ h_\sigma^{ij} \ln \left( \frac{h_\sigma^{ij}}{g_\gamma^{j}} \right) - h_\sigma^{ij} + g_\gamma^{j} \right] \, \mbox{is minimum.}
\label{eq_best_match_nexim}  	
\end{equation}

The derivative of this cost function with respect to prototypes $\vec{w}_j$ is explained in section~\ref{sec:xim_derivative}. It yields the following online learning update rule for a given data vector $\vec{x}$:	
\begin{equation}
 \vec{w}_j \leftarrow \vec{w}_j - \epsilon \, \Delta \vec{w}_j 
\label{eq_nexim_update}
\end{equation}	
\noindent with
\begin{equation}
	\Delta \vec{w}_j = \frac{\partial g_\gamma^{j}}{\partial \vec{w}_j} \left( 1 - \frac{h_\sigma^{ij}}{g_\gamma^{j}} \right).
\label{eq_nexim_update_delta}
\end{equation}
  
For the special case of a Gaussian neighborhood function in the high-dimensional data space $E$, i.e. $g_\gamma^{j} = \exp{\left(-\frac{d_E(\vec{x}, \vec{w}_j)}{2\gamma^2}\right)}, \; \gamma > 0$, the learning rule (\ref{eq_nexim_update_delta}) yields
\begin{equation}
		\Delta \vec{w}_j = \frac{1}{2 \gamma^2} \left( h_\sigma^{*j} - g_\gamma^{j} \right) \frac{\partial d_E(\vec{x}, \vec{w}_j)}{\partial \vec{w}_j},
\label{eq_nexim_update_delta_gauss}
\end{equation}
where $h_\sigma^{*j}$ may be chosen as a monotonically decreasing function, e.g.
\begin{itemize}
	\item [(i)] as a Gaussian
\begin{equation}
	h_\sigma^{*j} = \exp{\left(-\frac{d_O(\vec{r}^*,\vec{r}_j)}{2\sigma^2}\right)}\ \ \  \mbox{(XIM)},
\end{equation}	
	\item [(ii)] as a t-distribution	
\begin{equation}
h_\sigma^{*j} = \left( 1 + \frac{1}{\sigma} d_O(\vec{r}^*,\vec{r}_j) \right)^{-\frac{\sigma + 1}{2}}\ \ \  \mbox{(t-XIM)},
\end{equation}
	\item [(iii)] as a Cauchy-Lorentz distribution,	
\begin{equation}
h_\sigma^{*j} = \left( 1 + \frac{1}{\sigma^2} d_O(\vec{r}^*,\vec{r}_j) \right)^{-1}\ \ \  \mbox{(c-XIM)},
\end{equation}
\end{itemize}
where we call the resulting algorithms XIM, t-XIM, and c-XIM, respectively. Obviously (iii) can be seen as a special case of (ii).

For the special choice of a squared Euclidean metric in $E$, i.e.~$d_E(\vec{x}, \vec{w}_j) = \left( \vec{x} - \vec{w}_j \right)^2$, the learning rule (\ref{eq_nexim_update_delta_gauss}) yields
\begin{equation}
	\Delta \vec{w}_j = - \frac{1}{\gamma^2} \left( h_\sigma^{*j} - \exp{\left(-\frac{\left( \vec{x} - \vec{w}_j \right)^2}{2\gamma^2}\right)} \right) \left( \vec{x} - \vec{w}_j \right).
\label{eq_nexim_update_delta_euclid}	
\end{equation}


\section{Derivative of the XIM Cost Function}
\label{sec:xim_derivative}

With the abbreviation $h_{\sigma}^{ij} = h_\sigma(d_O(\vec{r}^i,\vec{r}^j))$ and $g_{\gamma}^{j} = g_\gamma(d_E(\vec{x}^i,\vec{w}^j))$, we write the derivative of the cost function equation (\ref{eq_cost_nexim}) with respect to a prototype vector $\vec{w}^k$:

\begin{eqnarray}
	\frac{\partial{E_{\mbox{\scriptsize XIM}}}}{\partial{\vec{w}^k}} & \propto & \int \sum_i \frac{\partial{\delta_{\Psi^{\mbox{\tiny GKL}}(\vec{x}),\vec{r}^i}}}{\partial{\vec{w}^k}} \sum_j \left[ h_\sigma^{ij} \ln \left( \frac{h_\sigma^{ij}}{g_\gamma^{j}} \right) - h_\sigma^{ij} + g_\gamma^{j} \right] p(\vec{x}) d\vec{x} \nonumber\\
&	+ & \int \sum_i \delta_{\Psi^{\mbox{\tiny GKL}}(\vec{x}),\vec{r}^i} \frac{\partial g_\gamma^{k}}{\partial \vec{w}^k} \left( 1 - \frac{h_\sigma^{ik}}{g_\gamma^{k}} \right) p(\vec{x}) d\vec{x},
\label{eq_derivative_cost_nexim}
\end{eqnarray}

\noindent with $\vec{r}^{* \mbox{\scriptsize GKL}}(\vec{x}) = \vec{r}^*(\vec{x})$ as defined in equation (\ref{eq_best_match_nexim}). The latter term yields the learning rule. The first term vanishes, as can be seen as follows: We use the shorthand notation

\begin{equation}
	\Phi^N(\vec{r}^i,\vec{x}) = \sum_j \left[ h_\sigma^{ij} \ln \left( \frac{h_\sigma^{ij}}{g_\gamma^{j}} \right) - h_\sigma^{ij} + g_\gamma^{j} \right]. 
\end{equation}

Then the best-match node can be expressed as

\begin{equation}
	\delta_{\vec{r}^{* \mbox{\tiny GKL}}(\vec{x}),\vec{r}^i} = H \left( \sum_k H(\Phi^N(\vec{r}^i,\vec{x}) - \Phi^N(\vec{r}^k,\vec{x})) - n + \frac{1}{2} \right).
\end{equation}

Hence the additional first term of equation (\ref{eq_derivative_cost_nexim}) vanishes, because of the following:

\begin{eqnarray}
\lefteqn{\int \sum_i \frac{\partial{\delta_{\vec{r}^{* \mbox{\tiny GKL}}(\vec{x}),\vec{r}^i}}}{\partial{\vec{w}^k}} \Phi^N(\vec{r}^i,\vec{x}) p(\vec{x}) d\vec{x}} \nonumber\\
& = & \int \sum_i \delta \left( \sum_l H(\Phi^N(\vec{r}^i,\vec{x}) - \Phi^N(\vec{r}^l,\vec{x})) - n + \frac{1}{2} \right) \nonumber\\
& & \cdot \sum_l \delta(\Phi^N(\vec{r}^i,\vec{x}) - \Phi^N(\vec{r}^l,\vec{x})) \cdot \left[ (g_{\gamma}^k - h_{\sigma}^{ik}) - (g_{\gamma}^k - h_{\sigma}^{lk}) \right] \frac{\partial g_\gamma^{k}}{\partial \vec{w}^k}
\end{eqnarray}

\begin{eqnarray}
\lefteqn{\sum_j \left[ h_\sigma^{ij} \ln \left( \frac{h_\sigma^{ij}}{g_\gamma^{j}} \right) - h_\sigma^{ij} + g_\gamma^{j} \right] p(\vec{x}) d\vec{x}} \nonumber\\
& = & \int \sum_{ilj} \delta \left( \sum_{l^{\prime}} H(\Phi^N(\vec{r}^i,\vec{x}) - \Phi^N(\vec{r}^{l^{\prime}},\vec{x})) - n + \frac{1}{2} \right) \nonumber\\
& & \cdot \, \delta(\Phi^N(\vec{r}^i,\vec{x}) - \Phi^N(\vec{r}^l,\vec{x})) \cdot (g_{\gamma}^k - h_{\sigma}^{ik}) \frac{\partial g_\gamma^{k}}{\partial \vec{w}^k} \nonumber\\
& & \cdot \left[ h_\sigma^{ij} \ln \left( \frac{h_\sigma^{ij}}{g_\gamma^{j}} \right) - h_\sigma^{ij} + g_\gamma^{j} \right] p(\vec{x}) d\vec{x} \nonumber\\
& & - \int \sum_{ilj} \delta \left( \sum_{l^{\prime}} H(\Phi^N(\vec{r}^i,\vec{x}) - \Phi^N(\vec{r}^{l^{\prime}},\vec{x})) - n + \frac{1}{2} \right) \nonumber\\
& & \cdot \, \delta(\Phi^N(\vec{r}^l,\vec{x}) - \Phi^N(\vec{r}^i,\vec{x})) \cdot (g_{\gamma}^k - h_{\sigma}^{ik}) \frac{\partial g_\gamma^{k}}{\partial \vec{w}^k} \nonumber\\
& & \cdot \left[ h_\sigma^{lj} \ln \left( \frac{h_\sigma^{lj}}{g_\gamma^{j}} \right) - h_\sigma^{lj} + g_\gamma^{j} \right] p(\vec{x}) d\vec{x}
\end{eqnarray}

\begin{eqnarray}
\lefteqn{\sum_j \left[ h_\sigma^{ij} \ln \left( \frac{h_\sigma^{ij}}{g_\gamma^{j}} \right) - h_\sigma^{ij} + g_\gamma^{j} \right] p(\vec{x}) d\vec{x}} \nonumber\\
& = & \int \sum_{il} \delta \left( \sum_{l^{\prime}} H(\Phi^N(\vec{r}^i,\vec{x}) - \Phi^N(\vec{r}^{l^{\prime}},\vec{x})) - n + \frac{1}{2} \right) \nonumber\\
& & \cdot \, \delta(\Phi^N(\vec{r}^i,\vec{x}) - \Phi^N(\vec{r}^l,\vec{x})) \cdot (g_{\gamma}^k - h_{\sigma}^{ik}) \frac{\partial g_\gamma^{k}}{\partial \vec{w}^k} \cdot \Phi^N(\vec{r}^i,\vec{x}) p(\vec{x}) d\vec{x} \nonumber\\
& & - \int \sum_{il} \delta \left( \sum_{l^{\prime}} H(\Phi^N(\vec{r}^l,\vec{x}) - \Phi^N(\vec{r}^{l^{\prime}},\vec{x})) - n + \frac{1}{2} \right) \nonumber\\
& & \cdot \, \delta(\Phi^N(\vec{r}^l,\vec{x}) - \Phi^N(\vec{r}^i,\vec{x})) \cdot (g_{\gamma}^k - h_{\sigma}^{ik}) \frac{\partial g_\gamma^{k}}{\partial \vec{w}^k} \cdot \Phi^N(\vec{r}^l,\vec{x}) p(\vec{x}) d\vec{x} = 0,
\end{eqnarray}

\noindent because of the symmetry of $\delta$ and the fact that $\delta$ is non-vanishing only if $\Phi^N(\vec{r}^l,\vec{x}) = \Phi^N(\vec{r}^i,\vec{x}))$.

\newpage

\section {Summary of the XIM Algorithm}
 
In summary, the XIM algorithm encompasses the following steps:
\\

\setlength{\fboxsep}{4mm}
\noindent \fbox{\parbox{0.96\textwidth}{
\begin{center}
\vspace{-0.2cm}
{\bf {\large Exploratory Inspection Machine (XIM)}}
\end{center}

{\bf Input:} Data vectors $\vec{x}$ or pairwise dissimilarities $d_E(\vec{x}_i, \vec{x}_k)$ (see section~\ref{sec:batch_xim}) 

\vspace{0.2cm}

{\bf Initialize:} Number of iterations $t_{\max}$, learning rate $\epsilon$, cooperativity parameters $\sigma$, $\gamma$, structure hypothesis $\vec{r}_j$, initial prototypes $\vec{w}_j$

\vspace{0.2cm}

{\bf For $t \leq t_{\max}$ do} 

\vspace{0.2cm}

{\bf begin} 

\vspace{-0.2cm}

\begin{enumerate}
\item Randomly draw a data vector $\vec{x}$.
\item Find best-match node using equation (\ref{eq_best_match_nexim}).
\item Compute neighborhood cooperativity $h_{\sigma}$ (may be pre-computed and retrieved from a look-up table).
\item Compute neighborhood cooperativity $g_{\gamma}$
\item Update prototypes according to equations (\ref{eq_nexim_update}) and (\ref{eq_nexim_update_delta}), e.g.~using equations (\ref{eq_nexim_update_delta_gauss}) -- (\ref{eq_nexim_update_delta_euclid}).
\end{enumerate}

\vspace{-0.2cm}

{\bf end}

\vspace{0.2cm}

{\bf Output:} Prototypes $\vec{w}_j$

\vspace{0.2cm}

{\bf Optional:} Compute explicit mapping by applying approximation or interpolation schemes to explicitly calculate a low-dimensional target point $\vec{y}$ for each data point $\vec{x}$, see section~\ref{sec:implementation}.

}} 

\vspace{0.1cm}
 
\section{Algorithmic Properties and Implementation Issues}
\label{sec:implementation}

XIM, unlike SNE and many other embedding algorithms, exhibits the interesting property that it allows to impose a prior structure on the projection space, which is a property that can also be found in XOM and SOM. Unlike SNE and many other data visualization techniques which exhibit a computational and memory complexity that is quadratic in the number of data points, the complexity of XIM can be easily controlled by the structure hypothesis definition, i.e.~the node grid in the observation space, and is linear with the number of data points and the number of prototypes. When implementing XIM, the following issues may be considered:

{\bf Best-match definition.} Although from a theoretical point of view, the best-match definition should be made according to equation (\ref{eq_best_match_nexim}), we found that using the much simpler minimal distance approach according to equation (\ref{eq_bestmatch_definition}) yields comparable results in practical computer simulations.

{\bf Weighting of attractive and repulsive forces.} It should be noted that the factor $\frac{1}{\gamma^2}$ in the update rules (\ref{eq_nexim_update_delta_gauss}) and (\ref{eq_nexim_update_delta_euclid}) can be seen as a quantity that determines the stepsize of a gradient descent step on the cost function, which can be treated as a free parameter similarly as the learning parameter $\epsilon$ in the online learning rule (\ref{eq_nexim_update}). We found it useful to omit the factor $\frac{1}{\gamma^2}$ and control the stepsize of gradient descent by annealing of the learning parameter $\epsilon$ only and to introduce a relative weighting of attractive and repulsive forces in learning rule (\ref{eq_nexim_update_delta_euclid}), i.e.

\begin{equation}
	\Delta \vec{w}_j = - \left( (1 - \eta) \, h_\sigma^{*j} - \eta \, \exp{\left(-\frac{\left( \vec{x} - \vec{w}_j \right)^2}{2\gamma^2}\right)} \right) \left( \vec{x} - \vec{w}_j \right),
\label{eq_nexim_update_delta_euclid_weighted}	
\end{equation}

\noindent where the method proved robust for a wide range of values for $\eta \in [0.1, 0.5]$.

Similarly as in other topographic mapping approaches, the parameters $\epsilon$, $\sigma$, and $\gamma$ may be adapted using annealing schemes, e.g.~by using an exponential decay

\begin{equation}
	\kappa(t) = \kappa(t_1) \, \left( \frac{\kappa(t_{\max})}{\kappa({t_1})} \right)^\frac{t}{t_{\max}}
\label{eq_annealing}
\end{equation}

\noindent where $\kappa := \epsilon$, $\kappa := \sigma$, or $\kappa := \gamma$, respectively, and $t \in [t_1, t_{\max}]$ denotes the iteration step.

As known from other algorithms, such as SNE or XOM, it can be useful to adapt the variance of the neighborhood cooperativity in the data space to the local `density' of neighbors, i.e.~to use a different $\gamma_i$ for each data sample $\vec{x}_i$ so that an $\alpha$-ball of variance $\gamma_i$ would contain a fixed number $k$ of neighbors. The rationale of this would be to ensure that data samples in less dense regions are adequately represented in the calculation of the embedding. Instead of annealing $\gamma$ directly, this number $k$ may be annealed, e.g. according to equation (\ref{eq_annealing}) by setting $\kappa := k$. An alternative approach is to find appropriate $\gamma_i$ by using the `perplexity' approach proposed for SNE in~\cite{Hinton2003}.

{\bf Explicit mapping of data to target points.} As the output of XIM is the location of prototypes $\vec{w}_j$ in the exploration space $E$, there is no explicit mapping of each data vector onto its target points. This marks a fundamental difference between XIM and XOM: As in XOM, each data vector has its own prototype, the final prototype positions already represent the low-dimensional embeddings for each data point. This does not hold for XIM. However, an explicit mapping can be accomplished easily by using the tuples $\left( \vec{w}_j, \vec{r}_j \right) \in E \times O$ as pairs of reference points for defining a mapping $\Psi$ which allows the user to find a low-dimensional target point $\vec{y}_i$ for each data point $\vec{x}_i$. Such an explicit mapping based on computed pairs of reference points can be calcualted by a wide range of interpolation or approximation procedures. Here, we specifically mention three approaches that have been successfully applied in similar contexts, namely Shepard's interpolation~\cite{Shepard68}, generalized radial basis functions softmax interpolation, e.g.~\cite{Axel2002x}, and the application of supervised learning methods. For the latter approach, $\left( \vec{w}_j, \vec{r}_j \right)$ pairs are used as labeled examples for models of supervised learning, such as feed-forward neural networks or other function approximators. For techical details of how to use these methods in a related context, we refer to our previous work~\cite{Axel2002x}. It should be noted that such explicit mapping approaches can also be used for `out-of-sample' extension, i.e.~for finding embeddings of new data points that have not previously been used for XIM training. 

\section{Batch XIM and XIM for Non-Metric Data}
\label{sec:batch_xim} 
Batch XIM algorithms can easily be derived from analyzing the stationary states of learning rules~(\ref{eq_nexim_update_delta}) -- (\ref{eq_nexim_update_delta_gauss}), where we expect the expectation values for these updates over all data points to be zero. For example, for learning rule~(\ref{eq_nexim_update_delta_gauss}), we expect the following condition to be fulfilled in the stationary state
\begin{equation}
	\mathcal{E}_i \left( f_{\vec{r}^*(\vec{x}_i), \vec{r}_j} \frac{\partial d_E(\vec{x}_i, \vec{w}_j)}{\partial \vec{w}_j} \right) = 0 \; \; \forall \; \; j \in \{1,\ldots,M\},
\end{equation}

\noindent where $\mathcal{E}_i$ denotes the expectation value over all data points $\vec{x}_i$ and $f_{\vec{r}^*(\vec{x}_i), \vec{r}_j} = h_\sigma^{*j} - g_\gamma^{j}$. For the squared Euclidean metric, we obtain the batch XIM fixed-point iteration
\begin{equation}
	\vec{w}_j = \frac{\sum_i f_{\vec{r}^*(\vec{x}_i), \vec{r}_j} \vec{x}_i} {\sum_i f_{\vec{r}^*(\vec{x}_i), \vec{r}_j}}.
\end{equation}
 
In analogy to equations (6) -- (8) in~\cite{Kohonen98}, even simpler and faster batch XIM methods can easily be obtained by introducing Voronoi sets of the exploration space according to the best-match criterion~(\ref{eq_best_match_nexim}). For practical implementations, however, the simple minimal distance criterion~(\ref{eq_bestmatch}) may be used.

Finally, we introduce {\bf XIM for non-metric data} by combining the batch XIM with the concept of the generalized median. This can be accomplished in full analogy to SOM for nonvectorial data according to~\cite{Kohonen2002}. However, a fundamental difference is that in forming the sum of distances for computing the generalized median, the contents of the node-specific data sublists within a neighborhood set of grid nodes should be weighted by the above cooperativity $f_{\vec{r}^*(\vec{x}_i), \vec{r}_j}$ rather than $h_\sigma^{*j}$ alone as in~\cite{Kohonen2002}. Likewise, the best-match definition~(\ref{eq_best_match_nexim}) should be used instead of a simple minimal distance criterion~(\ref{eq_bestmatch}). 

\section{Experiments}
\label{sec:realworld}

\begin{figure}[thb]
\vspace{-2.0cm}
  \parbox{0.49\textwidth}{
    \centerline{\psfig{figure=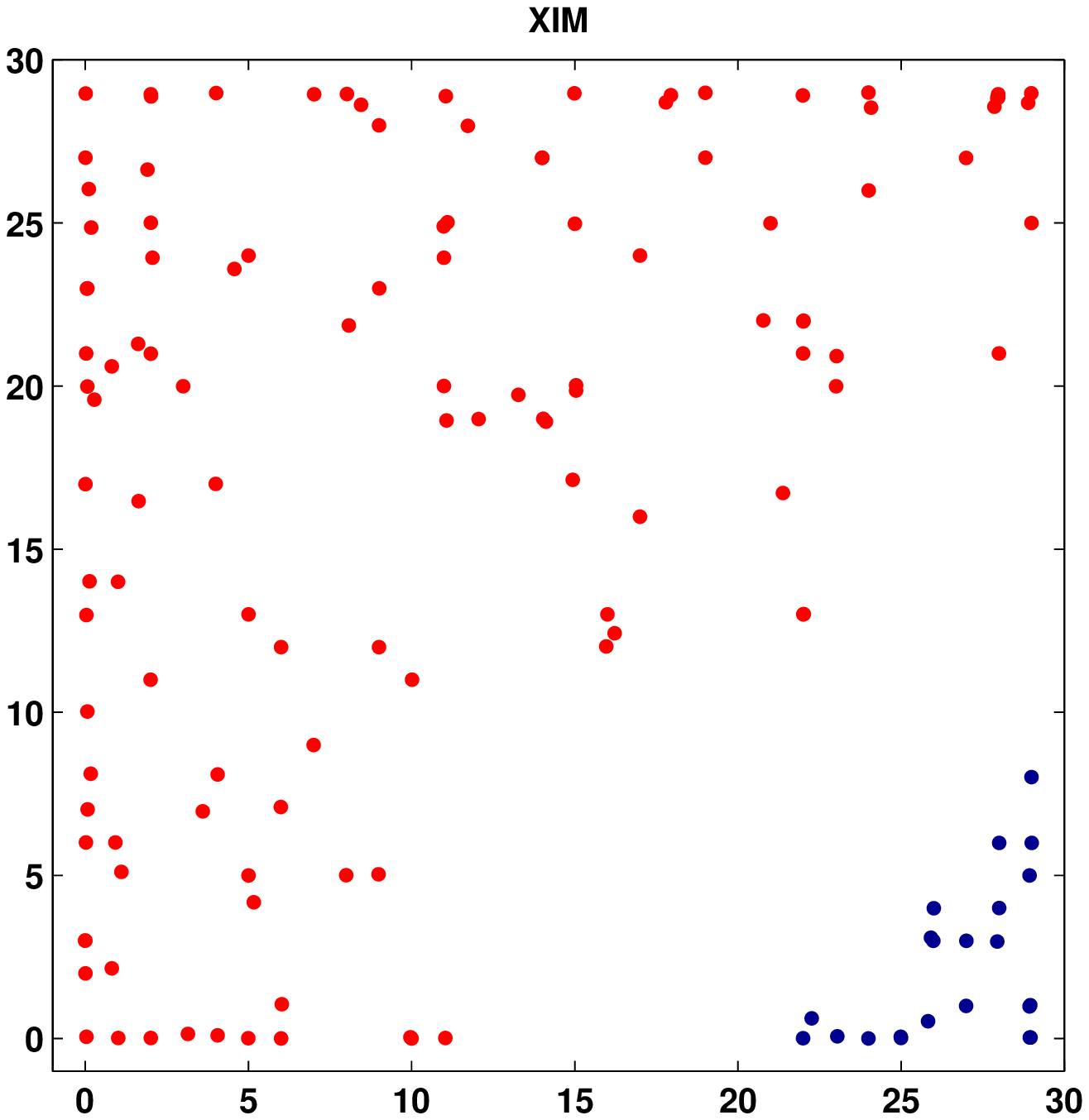,width=0.5\textwidth}}}
  \hfill \parbox{0.49\textwidth}{\centerline{\psfig{figure=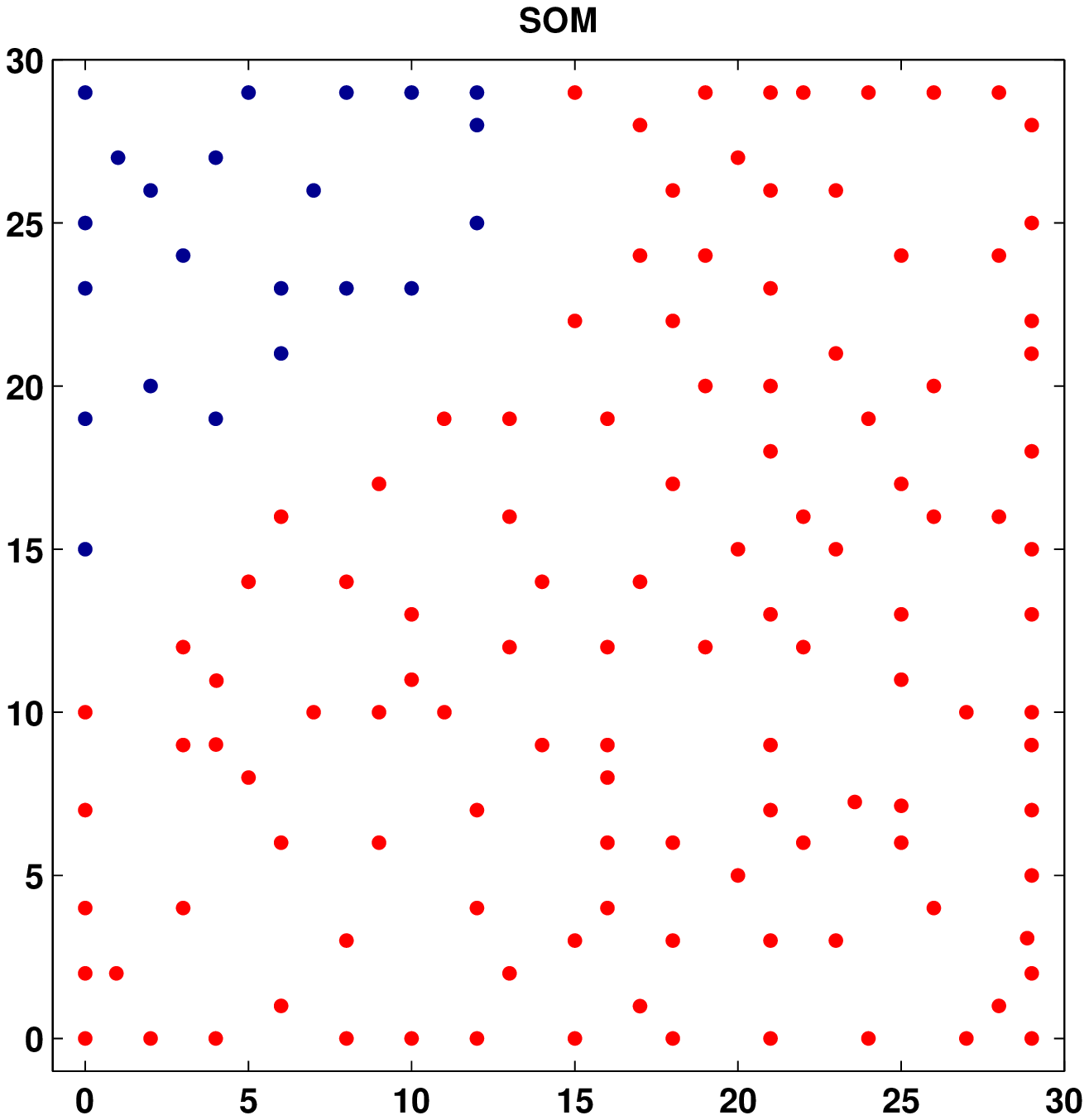,width=0.5\textwidth}}}
 \parbox{0.49\textwidth}{
 \vspace{0.2cm}
 \centerline{(A)}}
 \hspace{0.6cm}
 \parbox{0.49\textwidth}{
 \vspace{0.2cm}
 \centerline{(B)}}
%
\parbox{0.49\textwidth}{\centerline{\psfig{figure=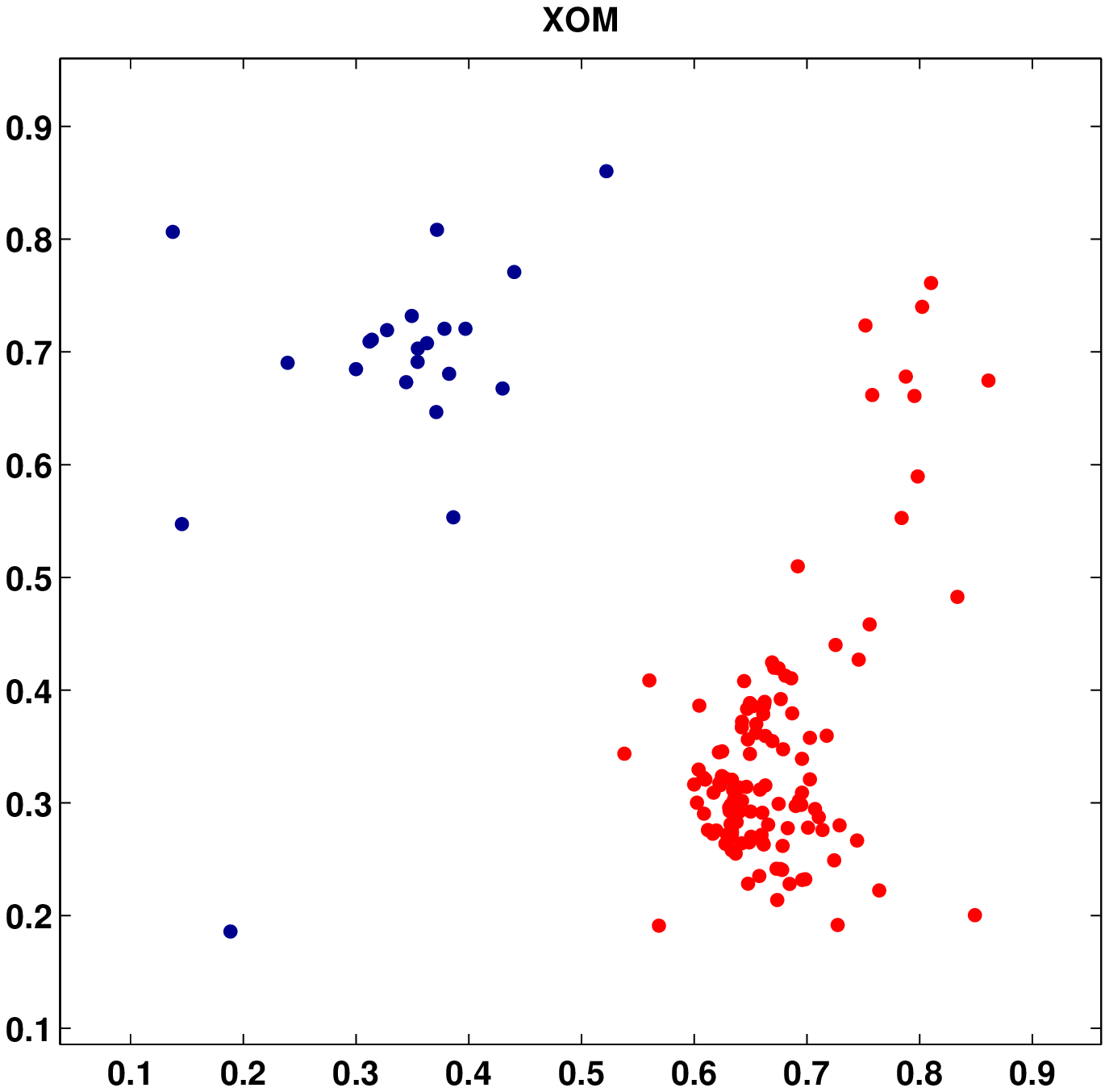,width=0.5\textwidth,height=0.51\textwidth}}}
  \hfill \parbox{0.49\textwidth}{\centerline{\psfig{figure=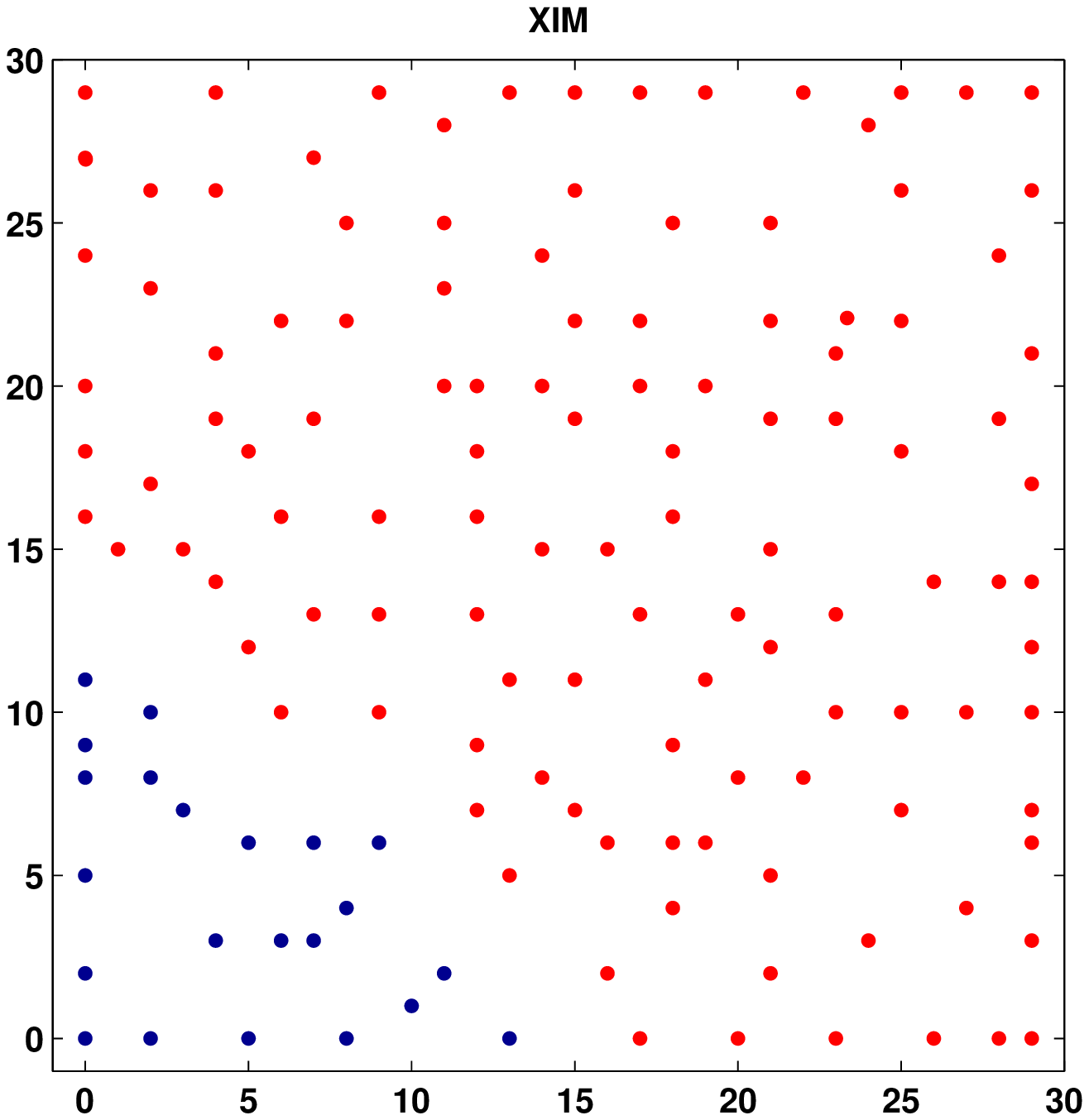,width=0.5\textwidth}}}
 \parbox{0.49\textwidth}{
 \vspace{0.2cm}
 \centerline{(C)}}
 \hspace{0.6cm}
 \parbox{0.49\textwidth}{
 \vspace{0.2cm}
 \centerline{(D)}}
\caption{Nonlinear dimensionality reduction of genome expression profiles related to ribosomal metabolism using (A) Cauchy-Lorentz Exploratory Inspection Machine (c-XIM), (B) Self-Organizing Map (SOM), (C) Exploratory Observation Machine (XOM), and (D) XIM with Gaussian neighborhood functions in the ordering space. For explanation, see text.} 
\label{fig_xim_ribosome}
\end{figure}

\begin{figure}[tbp]
\centerline{\psfig{figure=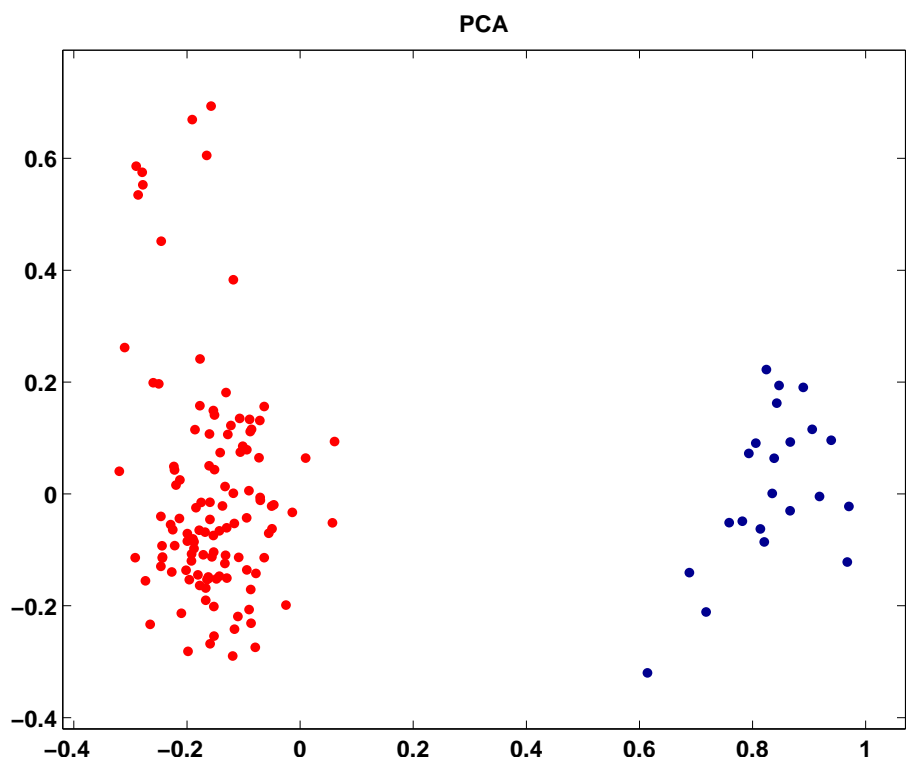,width=0.75\textwidth}}
\caption{Dimensionality reduction of genome expression profiles related to ribosomal metabolism using Principal Component Analysis (PCA). For explanation, see text.}
\label{fig_pca_ribosome}
\end{figure}

To prove the applicability of XIM, we present results on real-world data, see
Figs.~\ref{fig_xim_ribosome} and~\ref{fig_pca_ribosome}.  The data consists of 147 feature vectors in a 79-dimensional space encoding gene expression profiles obtained from
microarray experiments.


Fig.~\ref{fig_xim_ribosome} shows visualization results obtained from structure-preserving dimensionality reduction of gene expression profiles related to ribosomal metabolism, as a detailed visualization of a subset included in the genome-wide expression data taken from Eisen et al.~\cite{Eisen98}. The figure illustrates the exploratory analysis of the 147 genes labeled as `5' (22 genes) and `8' (125 genes) according to the cluster assignment by Eisen et al.~\cite{Eisen98}.  Besides several genes involved in respiration, cluster `5' (blue) contains genes
related to mitochondrial ribosomal metabolism, whereas cluster `8' (red) is
dominated by genes encoding ribosomal proteins and other proteins involved in
translation, such as initiation and elongation factors, and a tRNA synthetase.
The data has been described in~\cite{Axel_wsom_2009_tsp}.  

In the c-XIM genome map of Fig.~\ref{fig_xim_ribosome}A using a square grid of 30 $\times$ 30 nodes in the ordering space , it is clearly visible at
first glance that the data consists of two distinct clusters. Comparison with
the functional annotation known for these genes~\cite{Cherry97} reveals that
the map overtly separates expression profiles related to mitochondrial and to
extramitochondrial ribosomal metabolism. Fig.~\ref{fig_xim_ribosome}B shows a data representation obtained by a Self-Organizing Map (SOM) trained on the same data also using a square grid of 30 $\times$ 30 nodes in the ordering space. As can be clearly seen in the figure, SOM {\em cannot} achieve a satisfactory cluster separation in the mapping result as provided by c-XIM in Fig.~\ref{fig_xim_ribosome}A: Although the genes related to mitochondrial and to extramitochondrial ribosomal metabolism are collocated on the map, the distinct cluster structure underlying the data remains invisible, if the color coding is omitted. The result obtained by the Exploratory Observation Machine (XOM) in Fig.~\ref{fig_xim_ribosome}C as well as the mapping result obtained by Principal Component Analysis (PCA) in Fig.~\ref{fig_pca_ribosome}, however, can properly recover the underlying cluster structure. Note that the result obtained by XIM using Gaussian neighborhood functions in the ordering space (Fig.~\ref{fig_xim_ribosome}D) resembles the SOM result in that a clear cluster separation cannot be observed. The comparison of Figs.~\ref{fig_xim_ribosome}A and D thus demonstrate the benefit of introducing heavy-tailed neighborhood functions in the ordering space for this example.

A quantitative comparison of several quality measures for the results obtained in the data set of Figs.~\ref{fig_xim_ribosome} and~\ref{fig_pca_ribosome} is presented in~Tab.~\ref{tab_xim_ribosome}.


\newpage

\begin{table}[htb]
\caption{Comparative evaluation of quantitative embedding quality measures as known from the literature for the results obtained for dimensionality reduction of genome expression profiles related to ribosomal metabolism in Figs.~\ref{fig_xim_ribosome} and~\ref{fig_pca_ribosome}, namely Sammon's error~\cite{Sammon69}, Spearman's $\rho$,
as well as `trustworthiness' and `continuity' as defined by~\cite{Venna2007}. The table encompasses mean values and standard deviations obtained for 10 runs of each method, where in each run 140 out of 147 data points (approx.~95\% of the data) are randomly sub-sampled, i.e. embeddings and corresponding quality measures are computed for each of the 10 sub-sampled data sets. Trustworthiness and continuity values are computed as mean values obtained for $k = 1, \ldots, 50$ neighbors.
The free parameters of all the methods examined in the comparison (except PCA)
were optimized to obtain the best results with regard to the respective quality measure. Explicit mappings for c-XIM and SOM were computed using Shepard's interpolation~\cite{Shepard68}. Note that in comparison to SOM, c-XIM yields competitive results for several quality measures.}
\vspace{0.5cm}
\begin{center}
\begin{tabular}{|l||c|c|c|c|}
\hline
Method & Sammon & Spearman's $\rho$ & Trustworthiness & Continuity \\ 
\hline \hline
SOM & 0.18 (0.02) & 0.50 (0.05) & 0.84 (0.02) & 0.85 (0.01) \\
\hline
c-XIM & 0.17 (0.02) & 0.59 (0.09) & 0.87 (0.02) & 0.86 (0.02) \\
\hline
XOM & 0.19 (0.03) & 0.88 (0.02) & 0.86 (0.02) & 0.90 (0.02) \\
\hline
PCA & 0.18 (0.01) & 0.86 (0.01) & 0.85 (0.01) & 0.89 (0.01) \\ 
\hline
\end{tabular}
\clearpage
\label{tab_xim_ribosome}
\end{center}
\end{table}


\section{XIM Using Other Divergences}

XIM, as described so far, has been introduced based on the generalized Kullback-Leibler divergence. However, it is important to emphasize that XIM is not restricted to this specific divergence measure alone, but other divergences can be used easily. 
However, our work presented here differs substantially from prior approaches to utilize divergences for nonlinear dimensionality reduction. In contrast to SNE~\cite{Hinton2003} and t-SNE~\cite{van_der_Maaten2008}, divergence-based cost functions for nonlinear dimensionality reduction are {\em not} optimized {\em directly}, but via iterative incremental online learning. In contrast to~\cite{Villmann2011}, we do not use divergences as distance measures {\em within} the data or the embedding space, but as a dissimilarity measure {\em between} these spaces. In contrast to our previous work~\cite{Bunte2011}, 
which applies divergences to XOM, the XIM method presented in this paper systematically inverts the quoted approaches by introducing divergences in order to derive novel topographic vector quantization learning rules. This is accomplished by calculating the derivatives of divergence-based topographic vector quantization cost functions {\em with respect to distance measures in the high-dimensional data space} rather than derivatives {\em with respect to distance measures in the the low-dimensional embedding space}, such as in~\cite{Bunte2011} or~\cite{Bunte2011a}. 

As an important consequence, we can thus introduce successful approaches to alleviate the so-called 'crowding phenomenon' as described by~\cite{van_der_Maaten2008} into classical topographic vector quantization schemes, in which, in contrast to XOM and its variants, data items are associated with the {\em exploration} space and not with the {\em ordering} space of topographic mappings. Specifically, the deep difference when compared to the XOM approach becomes evident, when heavy-tailed distributions in the low-dimensional embedding space are utilized to address this issue: The update rules for the resulting t-XIM (and c-XIM) algorithms are significantly different from the t-NE-XOM update rules as specified in~\cite{Bunte2011} for the Kullback-Leibler divergence and in~\cite{Bunte2011a} for other divergences. The reason is that for t-XIM and c-XIM, in contrast to NE-XOM and its variants, there is no need to re-compute the derivative of the neighborhood function $h$ of the low-dimensional embedding space with respect to its underlying distance measure when computing the new learning rules. Instead, only $h$, not $g$ has to be replaced by a heavy-tailed distribution. In other words, we can derive learning rules for XIM by directly adopting derivatives of divergences as known from the mathematical literature and creatively applying them to a completely different context.

\section{Extensions of XIM}

XIM is likely to be even more useful by slight extensions or in combination with other methods of data analysis and statistical learning. The combination with interpolation and approximation schemes for explicit mapping between data and embedding spaces has already been discussed above. Other straightforward variants of XIM comprise the iterative use of XOM, local or hierarchical data processing schemes, e.g.~by initial vector quantization, and variable choices of 'batch', 'growing', or 'speed-up' variants analogous to those described in the literature on topology-preserving mappings. 

As noted earlier, there are no principal restrictions to input data and structure hypotheses. For example, they may be subject to a non-Euclidean, e.g. hyperbolic, geometry, or even represent nonmetric dissimilarities. Distances in the observation and exploration spaces may be rescaled dynamically during XIM training. These slight extensions should further enhance the applicability of XIM to many areas of information processing.  

\section*{Acknowledgements}
The author thanks Markus B. Huber and Mahesh B. Nagarjan for performing computer simulations on XIM. 


\end {document}